\documentclass[10pt]{article}

\usepackage{amsmath}
\usepackage{amssymb}
\usepackage{graphicx}
\usepackage{cite}
\usepackage{color}

\usepackage[labelfont=bf,labelsep=period,justification=raggedright]{caption}

\makeatletter
\renewcommand{\@biblabel}[1]{\quad#1.}
\makeatother

\date{}

\newcommand{\eg}{\textit{e.g.} }
\newcommand{\ie}{\textit{i.e.} }

\begin{document}

\begin{flushleft}
{\Large
\textbf{Multi-environment model estimation for motility analysis of {\it Caenorhabditis Elegans}}
}
\\
Raphael Sznitman$^{1}$,
Manaswi Gupta$^{1}$,
Gregory D. Hager$^{1}$,
Paulo E. Arratia$^{2}$,
and Josu{\'e} Sznitman$^{3,\ast}$
\\
\bf{1} Department of Computer Science, Johns Hopkins University, Baltimore MD 21218, USA
\\
\bf{2} Department of Mechanical Engineering and Applied Mechanics, University of Pennsylvania, Philadelphia PA 19104, USA
\\
\bf{3} Department of Mechanical and Aerospace Engineering, Princeton University, Princeton NJ 08544, USA
\\
$\ast$ E-mail: sznitman@princeton.edu
\end{flushleft}

\noindent
\textbf{Running title:} Motility analysis of {\it C. elegans} with MEME

\section*{Abstract}
The nematode {\it Caenorhabditis elegans} is a well-known model organism used to investigate fundamental questions in biology. Motility assays of this small roundworm are designed to study the relationships between genes and behavior. Commonly, motility analysis is used to classify nematode movements and characterize them quantitatively. Over the past years, {\it C. elegans}' motility has been studied across a wide range of environments, including crawling on substrates, swimming in fluids, and locomoting through microfluidic substrates. However, each environment often requires customized image processing tools relying on heuristic parameter tuning. In the present study, we propose a novel Multi-Environment Model Estimation (MEME) framework for automated image segmentation that is versatile across various environments. The MEME platform is constructed around the concept of Mixture of Gaussian (MOG) models, where statistical models for both the background environment and the nematode appearance are explicitly learned and used to accurately segment a target nematode. Our method is designed to simplify the burden often imposed on users; here, only a \textit{single} image which includes a nematode in its environment must be provided for model learning. In addition, our platform enables the extraction of nematode `skeletons' for straightforward motility quantification. We test our algorithm on various locomotive environments and compare performances with an intensity-based thresholding method. Overall, MEME outperforms the threshold-based approach for the overwhelming majority of cases examined. Ultimately, MEME provides researchers with an attractive platform for {\it C. elegans}' segmentation and `skeletonizing' across a wide range of motility assays.

\section*{Introduction}
Since its introduction in the laboratory over thirty years ago \cite{brenner1974}, the nematode {\it Caenorhabditis elegans} has become a ubiquitous model organism to study fundamental questions in biology \cite{rankin2002}. In particular, {\it C. elegans} is now widely used as a platform for drug screening and development \cite{link2000,jorgensen2002}, as well as for modeling various aspects of human diseases \cite{chamberlain2000,silverman2009}. In the quest to understand the relationships between genes and behavior, this small, approximately 1~mm long roundworm offers a number of advantages for laboratory applications. These include a short life cycle, the availability of many mutants to explore gene functions, knowledge of its complete cell lineage \cite{sulston1977,sulston1983}, simplicity of the nervous system and its wiring \cite{white1986}, and a fully sequenced genome \cite{consortium1998}.

\begin{figure}[!ht]
\centering
\includegraphics[width=1.0\textwidth]{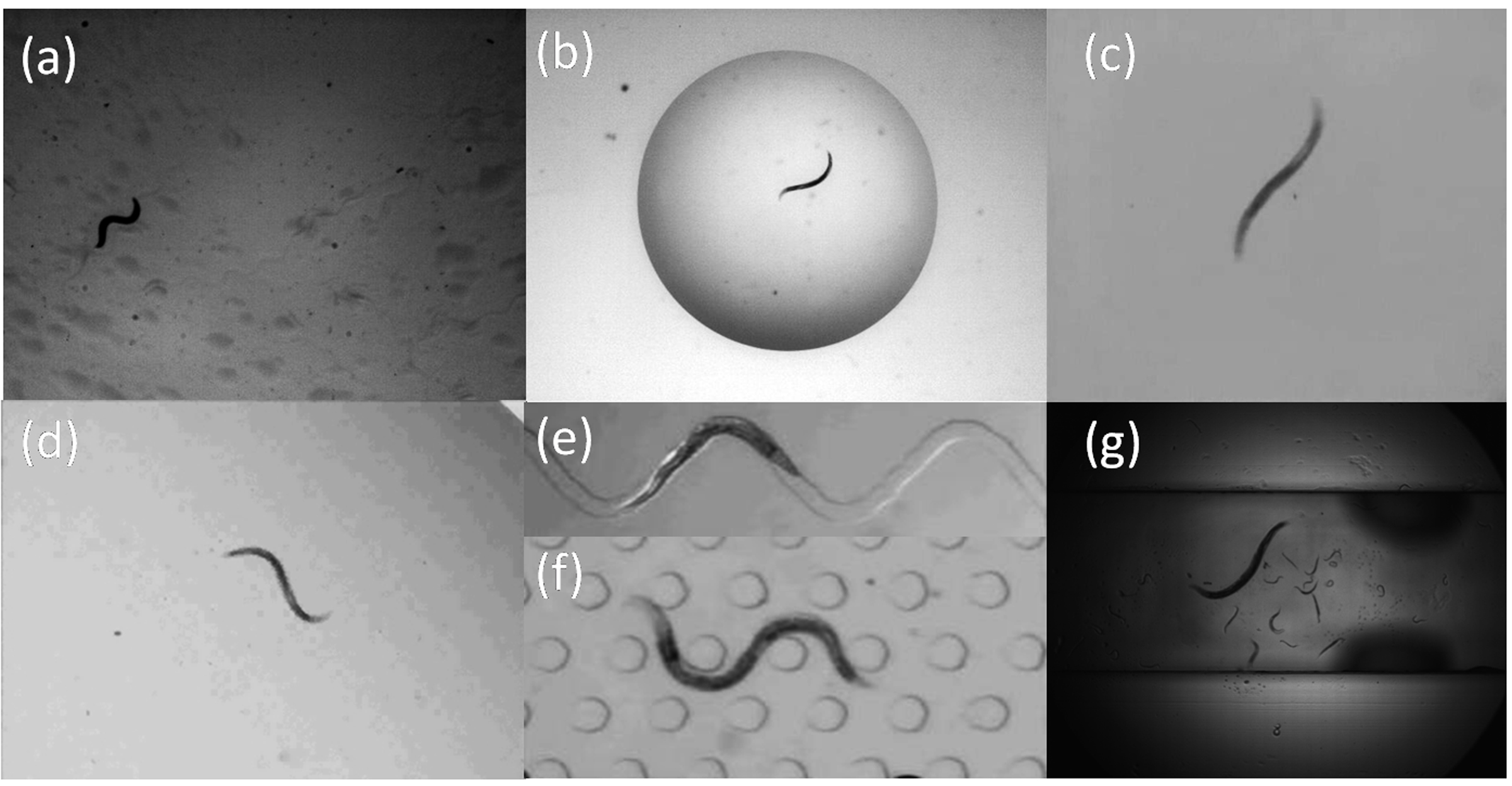}
\caption{{\bf Examples of environments used in {\it C. elegans}' motility assays.} {\bf (a)} Nematode crawling on an agar plate (Supplementary Video S2). {\bf (b)} Nematode swimming in a 5~ml drop of M9 buffer solution (Supplementary Video S4). {\bf (c)} Nematode swimming in a solution of gelatin dissolved in M9 (source: berrigel0.0perc.mov, Supplementary Material in Berri {\it et al.} \cite{berri2009}). {\bf (d)} Nematode swimming inside a fluid-filled chamber (source: SM2.avi, Supplementary Material in Pierce-Shimomura {\it et al.} \cite{pierce2008}). {\bf (e)}-{\bf (f)} Nematode locomotion in a microfluidic substrate (source: Supplemental Videos 2 and 4 in Lockery {\it et al.} \cite{lockery2008}). {\bf (g)} Nematode swimming in a shallow acrylic channel filled with M9 (Supplementary Video S5). Nematodes shown in (a) through (g) are wild-type (N2) {\it C. elegans} and are all approximately $1~$mm long.
}
\label{Fig_1}
\end{figure}

A widespread strategy to investigate the genetic basis of behavior is to classify nematode movements and characterize them quantitatively. Traditionally, motility quantification has been based on crawling assays \cite{karbow2006,pierce1999,pierce2008,tavernarakis1997}, where {\it C. elegans} is observed to crawl on a substrate (\eg agar plate). This is shown for example in Fig.~\ref{Fig_1}(a). In the recent past, however, the number of environments used for nematode motility assays has vastly expanded. Studies of {\it C. elegans}' motility behavior now include various swimming assays \cite{berri2009,korta2007,ghosh2008,pierce2008,sznitman2010,sznitman2010b}, as shown in Fig.~\ref{Fig_1}(b)-(d) and (g). In parallel, with the widespread availability of microfabrication techniques, nematode motility assays are increasingly conducted in microfluidic environments \cite{chronis2007,lockery2008,park2008,qin2007}. An example of such environments is shown in Fig.~\ref{Fig_1}(e) and (f). This latter platform has become particularly attractive for high-throughput drug screening applications \cite{rhode2007,shi2008}. Overall, with the growing variations in environments used for nematode behavioral assays, users are in need of reliable image analysis tools capable of extracting quantitative data across a wide spectrum of experimental mediums.

The analysis of motility behavior has traditionally relied on qualitative observations to describe {\it C. elegans}' locomotion and discriminate between wild-type and mutant nematodes. In many instances, however, qualitative variations between strains are not apparent to the trained eye (as in \cite{bessou1998}). Such limitations have sprouted the development of automated image analysis systems in an effort to deliver relevant phenotypic differences between nematode strains \cite{baek2002,buckingham2008,buckingham2009,cronin2005,feng2004,geng2004,huang2006,huang2008,hoshi2006,tsibidis2007}. While the bulk of the research effort has been directed at analyzing locomotive traits of individual nematodes, some multi-worm tracking and feature extraction systems have also been developed \cite{ramot2008,roussel2007,tsechpenakis2008}. Yet, the majority of state-of-the-art image analysis systems are designed for a {\it specific} environment, most commonly crawling \cite{baek2002,cronin2005,geng2004,feng2004,huang2006,huang2008,ramot2008} or swimming assays \cite{buckingham2009,tsechpenakis2008}. These systems induce a tradeoff between either limiting the range of possible assays a researcher will investigate for motility analysis or customizing segmentation parameter selection across varying environments. However, an optimal system for the user is one which ideally bypasses such compromise.

\begin{figure}[!ht]
\centering
\includegraphics[width=1.0\textwidth]{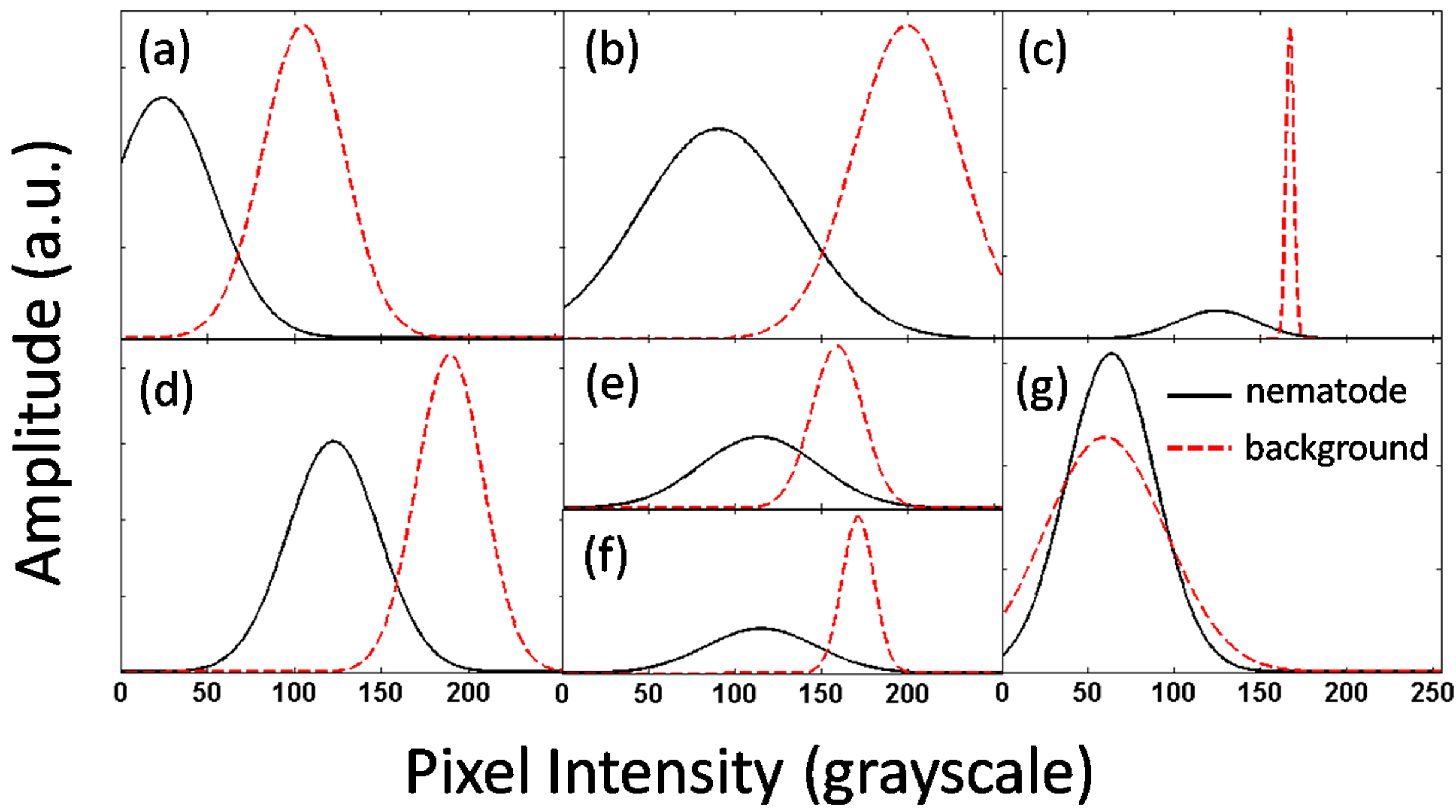}
\caption{{\bf Pixel intensity distributions of nematode and background environment}. Plots {\bf (a)} through {\bf (g)} correspond to distributions obtained from the images of Fig.~\ref{Fig_1}. Grayscale pixel intensities vary between 0 (black) and 255 (white). Distributions are assumed to be Gaussian and parametrized with the mean and standard deviation of pixel intensities.}
\label{Fig_2}
\end{figure}

Current image analysis systems provide users with morphological and locomotion features to quantify behavioral phenotypes of {\it C. elegans}. Such features include amongst other nematode speed \cite{ramot2008,tsibidis2007}, wavelength and frequency of body undulations \cite{buckingham2009,cronin2005}, body curvature \cite{baek2002,feng2004}, and omega bends \cite{huang2006}; several of which make use of nematode centerline data, also known as `skeletons'. In practice, features are extracted from binary images, or {\it segmentations}, separating the nematode from its environment, or {\it background}. Several analysis systems compute binary images by applying a simple intensity-based threshold at each pixel location \cite{geng2003, feng2004, cronin2005, hoshi2006, tsibidis2007, huang2009}. Most commonly, this involves having the user manually select an appropriate range of intensities which characterizes the nematode. A variation to this approach has been the use of an adaptive threshold where nematode intensities, or {\it appearance}, are assumed to significantly differ from the average background intensities \cite{huang2008, huang2006, baek2002}. While these methods have shown promising capabilities, the range of environments for which they can be used for is in fact limited. This limitation is illustrated in Fig.~\ref{Fig_2} where the pixel intensity distributions of the nematode and background are respectively plotted for the environments shown in Fig.~\ref{Fig_1}. Here, distributions are assumed to be Gaussian and parametrized with the mean and standard deviation of pixel intensities. In none of the cases shown can a single threshold separate any pair of distributions without causing significant errors (Fig.~\ref{Fig_2}). While threshold-based techniques can still be used to compute accurate segmentations, this requires significant effort on the user-end to adjust appropriate threshold values along with other noise canceling schemes (\eg median filters, morphological operators, background subtraction, etc.). Altogether, this tedious process makes thresholding ill-suited for applications in complex background environments.

More recently, alternative approaches to thresholding techniques have been pursued. For example, the work of Stauffer and Grimson \cite{stauffer1999cvpr} has been applied to the problem of nematode segmentation \cite{roussel2007, tsechpenakis2008}. Here, the idea is to systematically learn how background pixels are individually distributed and use this information to segment the nematode. The learning process is done using a set of \textit{training} images to statistically model the appearance of the background. Namely, each pixel is modeled by means of a Mixture of Gaussians (MOG), where the parameters of the model are learned from the training image set. This approach has been recently shown to provide excellent results for nematode segmentation \cite{roussel2007, tsechpenakis2008} as well as for other applications \cite{stauffer1999cvpr, piccardi2004csmc, sznitman2009iccv}. A major drawback, however, of modeling pixels with MOGs is that many parameters must be learned; this requires a large set of nematode-free training images. This condition largely prohibits extracting nematode segmentations from arbitrary sequences (\eg open-access material).

In the present study, we propose a novel framework for image analysis of {\it C. elegans} that is versatile across a wide range of environments. Moreover, our system is designed to greatly simplify the burden imposed on the user end; only a \textit{single} image from a sequence which includes a nematode in its environment must be provided. From this input, models for {\it both} the background and the nematode appearance are individually learned using MOGs (see Methods). These models are then applied to segment the nematode in subsequent images. Next, we provide an original algorithm for extracting nematode skeletons for applications to {\it C. elegans}' behavioral assays. Nematode segmentation and skeletonizing algorithms have been packaged together in a software for straightforward use by a broad range of researchers. We test our image analysis system on representative locomotive environments (Fig.~\ref{Fig_1}) and compare performances with a state-of-the-art thresholding method (see Results). Finally, we illustrate some examples of motility metrics (\eg body curvature) which are commonly sought from nematode skeleton data (see Discussion). We discuss future directions for algorithmic improvement (\eg multi-worm tracking) as well as new directions for potential applications (\eg cell tracking).

\section*{Methods}
\label{sec:meme}

\begin{figure}
\centering
\includegraphics[width=0.95\textwidth]{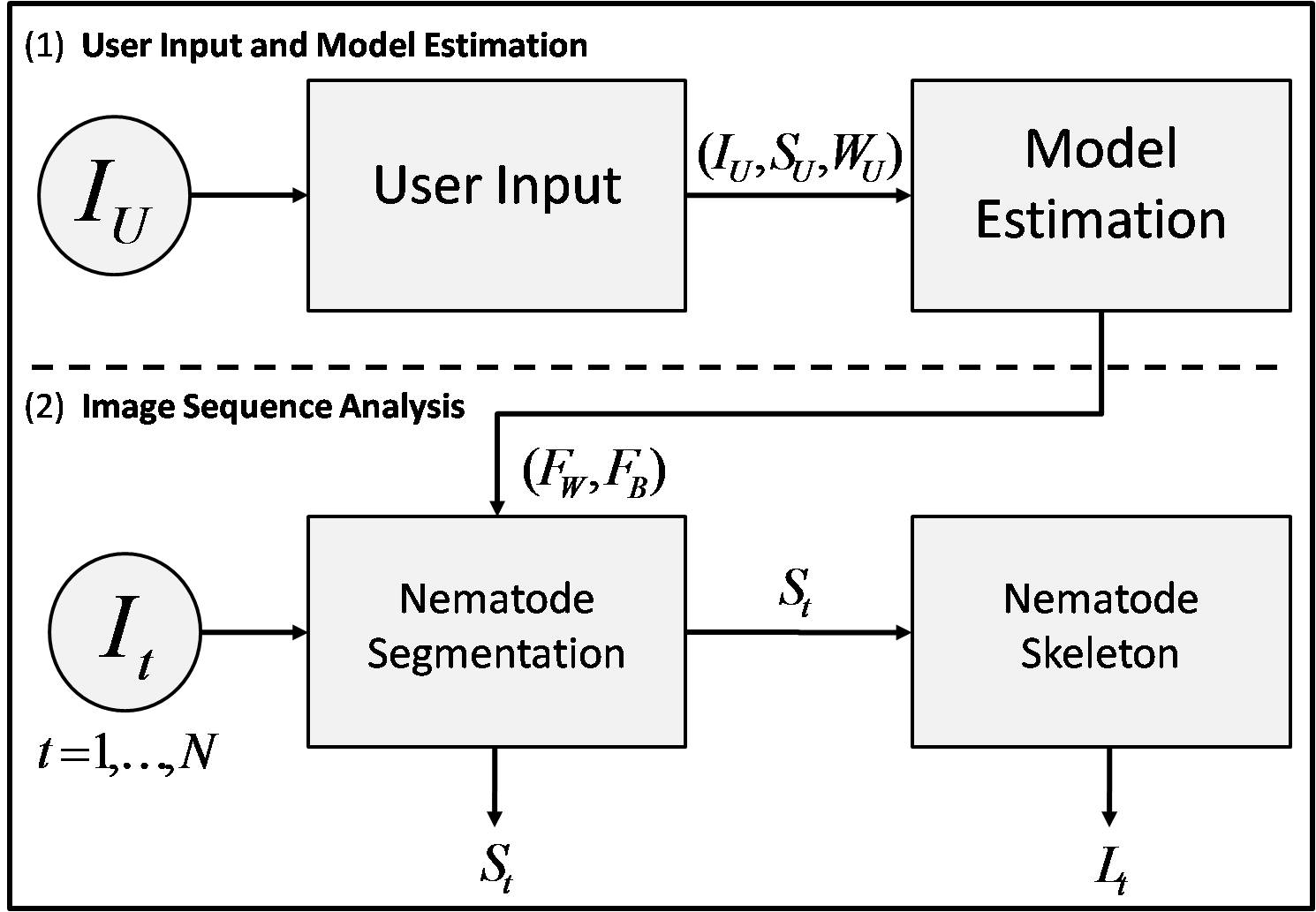}
\caption{
{\bf Multi-Environment Model Estimation (MEME) framework overview.} The system consists of two components: (1) a user input and learning stage and (2) an image analysis stage. In stage (1), the user provides an image ($I_U$) and marks the nematode boundary ($S_U$) and width ($W_U$). From this input, appearance models for the nematode ($F_W$) and background ($F_B$) are learned. In (2), for each image ($I_t$) in a sequence, nematodes are then segmented ($S_t$) by using $F_W$ and $F_B$. Nematode skeletons ($L_t$) are then extracted from these segmentations.
}
\label{Fig_3}
\end{figure}

The \textit{Multi-Environment Model Estimation} (MEME) framework consists of two sequential stages. As a first step, (i) the user provides information, allowing a model for the nematode and the background environment to be learned. In the second stage, (ii) the nematode and background models are used to segment the nematode and then extract its skeleton for a sequence of images. In stage (i), the user is required to input a hand-segmentation of the nematode and its corresponding width for a \textit{single} image (see Supplementary Video S1). This approach can be viewed as a form of ``One Shot Learning" \cite{fleuret05,feifei06}, where model learning occurs only once, from a single labeled example. A flowchart of the MEME framework is schematically shown in Fig.~\ref{Fig_3}.

We briefly introduce the notation used throughout the article. We define the sequence of images provided by the user as $\mathcal{I}= \{I_1,\ldots,I_N\}$ for $N$ discrete time steps, where at each time step $t$, $I_t \in [0,255]^{n \times m}$; $n$ and $m$ are the width and height of each image, respectively. We denote the user provided data ($U$) as a triple $U = \{I_U,S_U,W_U\}$, where $I_U$ is an image from $\mathcal{I}$, $S_U$ is the nematode body segmentation and $W_U$ is the nematode width (Fig.~\ref{Fig_3}). We specify the intensity models derived from $U$ as $F_W$ for the worm and $F_B$ for the background. For any given image $I_t$ included in $\mathcal{I}$, we define the computed segmentation of the nematode as $S_t$ and the list of ordered pixel coordinates describing the nematode skeleton as $L_t$.

\subsection*{Nematode Segmentation}
\label{sec:meme_segmentation}


Our first step is to provide an automatic mechanism to compute accurate nematode segmentations for a single target using a static camera (as in \cite{roussel2007, tsechpenakis2008}). To do this, we build on the idea of using Mixtures of Gaussians (MOG) \cite{stauffer1999cvpr,piccardi2004csmc,sznitman2009iccv} to construct accurate and robust appearance models for the background and the nematode. As it is often the case for MOG methods, each pixel in an image is treated as a random variable which can be modeled by summing $K$ weighted Gaussian distributions. This can be formally written as

\begin{equation}
P(x) = \sum_{i=1}^{K} \pi_i \mathcal{G}(x | \mu_i, \Sigma_i),
\end{equation}

\noindent
where $x$ is a pixel intensity value, $\pi_i, \mu_i$ and $\Sigma_i$ are respectively the weight, mean and covariance of the $i$th Gaussian distribution $\mathcal{G}$. These parameters are usually estimated by means of an Expectation-Maximization (EM) algorithm \cite{dempster77}. The intuition behind this model is that each individual Gaussian represents the appearance a particular pixel may take. Therefore, combining each Gaussian provides a way to model complex pixel observations. Typically, doing this over all pixels in an image is an effective way to model background scenes \cite{stauffer1999cvpr,piccardi2004csmc,sznitman2009iccv}.

A consequence of this approach is that the number of MOGs used is considerable (\ie the total number $nm$ of pixels in an image) and the number of parameters required is very large $(3Knm)$. In turn, a substantial number of images is needed to estimate the MOG parameters accurately as each image only provides a single sample for each MOG (see \cite{stauffer1999cvpr} for more details). Moreover, the entire background scene must be visible when attempting to estimate these parameters, since each image is used to model the background and not the nematode. This latter condition becomes problematic when image sequences always contain a nematode in the field of view (\eg Supplementary Videos available in references \cite{berri2009,lockery2008,pierce2008}).

\begin{figure}[t]
\centering
\includegraphics[width=0.95\textwidth]{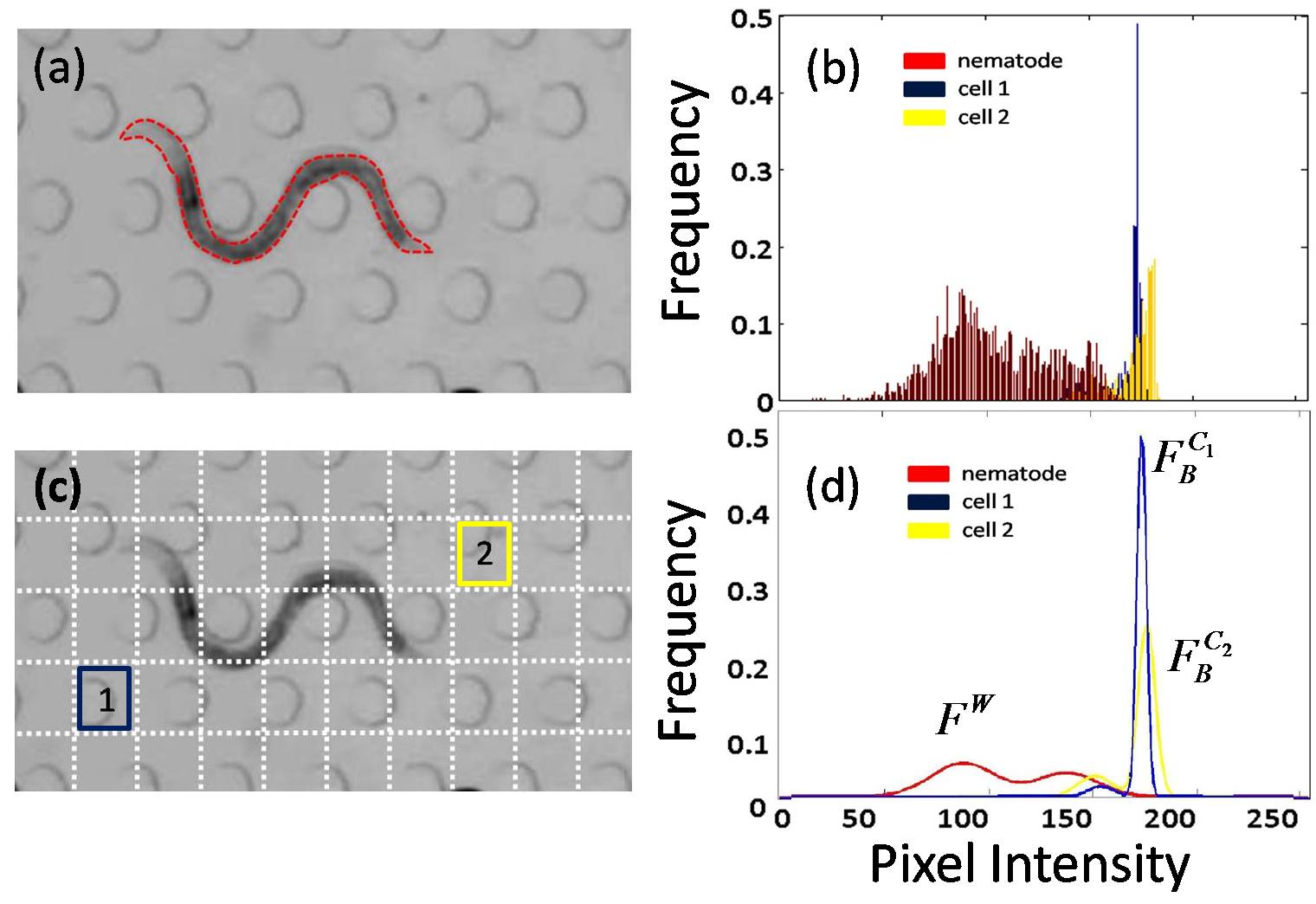}
\caption{
{\bf User input and nematode segmentation.} Figures {\bf(a)} through {\bf(d)} illustrate the stages of the segmentation process in MEME for a sample environment (source: Supplemental Video in Lockery {\it et al.} \cite{lockery2008}). {\bf(a)} The user selects the boundary of the nematode on a given image. From this manual segmentation, the distribution of nematode features can be extracted. {\bf(b)} Distribution of pixel intensities from the nematode region. Here, $d=1$ for illustrative purposes.  {\bf(c)} The background scene is partitioned into a grid, where each cell corresponds to a particular pixel block. Two arbitrary cells are labeled for clarity; their corresponding intensity distributions are shown in {\bf(b)}. For both the nematode and the cells, MOG parameters are then learned. {\bf(d)} Representation of the MOG models for the nematode ($F^W$) and the two background cells ($F^{C_1}_B$ and $F^{C_2}_B$).}
\label{Fig_4}
\end{figure}

To avoid such drawbacks, we choose instead to model the nematode appearance ($F_W$) in addition to the background model ($F_B$) by means of MOGs. To learn the parameters of $F_W$, we use the information gathered from the user. Namely, $S_U$ provides the pixel region of $I_U$ containing the nematode as delineated by the user (Fig.~\ref{Fig_4}a). From this region, we randomly select pixel locations and extract $[d \times d]$ image patches from $I_U$ around each location. These patches are then vectorized and treated as independent samples. We denote this feature extract process as $f(u,v;I,d) = x \in \mathbb{R}^{d^2}$, where we select a patch around pixels $(u,v)$ for any given image $I$. Notice that when $d = 1$, this reduces to sampling the selected pixels only; Fig.~\ref{Fig_4}(b) shows the histogram of intensities for the case $d=1$. In general, applying this transformation allows for modeling intensities with respect to image patches, as this approach carries more information than individual and independent pixels. Computing an appropriate value for $d$ is done by using a linear model (\ie $d = \alpha_{1}\frac{W_U}{max(m,n)} + \alpha_0$, where $\alpha_1$ and $\alpha_0$ are constants). The samples extracted are then used to estimate the parameters of $F_W$ by using the EM algorithm. Figure~\ref{Fig_4}(d) illustrates a visual representation of the estimated MOGs of $F_W$ for $K=2$ and $d=1$.

Next, modeling of the background ($F_B$) is done by partitioning the image ($I_U$) into $\tilde{n}$ distinct and non-overlapping cells, $\mathcal{C} = \{C_{1},\ldots,C_{\tilde{n}} \}$, as shown in Fig.~\ref{Fig_4}(c). Each cell ($C_{c}$) is then treated as a random variable and modeled with its own independent MOG. Hence, $F_B = \{F^{C_{1}}_B,\ldots,F^{C_{\tilde{n}}}_B\}$, where each pixel in $I_t$ is associated with a unique $F^{C_c}_B$; in our implementation, we choose $\tilde{n}=10\times10$. The parameters of each $F^{C_c}_B$ are computed from extracted samples in the partition $C_{c}$. Similarly to building the nematode model, samples are $[d\times d]$ pixel patches from $I_U$, which have been vectorized. Examples of intensity distributions for two arbitrary cells (Fig.~\ref{Fig_4}c) are displayed in Fig.~\ref{Fig_4}(b) along with their corresponding MOG representations in Fig.~\ref{Fig_4}(d). Two consequences arise from such partitioning. First, only a total of $3K\tilde{n}$ parameters need to be estimated, as opposed to $3Knm$. Secondly, a single image is sufficient to estimate these parameters, as background regions covered by the nematode can still be modeled by neighboring pixels in a cell. This reduces the number of training images required and relaxes the constraint that training images must only contain background pixels. Note that when a cell is reduced to a single pixel ($\tilde{n} \approx nm$), $F_B$ is similar to the model described in \cite{roussel2007, tsechpenakis2008}. Alternatively, when a cell corresponds to the entire image ($\tilde{n} = 1$), $F_B$ is similar to models typically used by thresholding techniques \cite{huang2008, huang2006, baek2002}.

Nematode segmentation for image $I_t$ can then be computed at each pixel $(u,v)$, belonging to cell $C_c$, as

\begin{equation}
\label{eq:homogeneity}
S_t(u,v) = \left\{
\begin{array}{ll}
1 & \textrm{ if } r( f(u,v;I_t,d), c) > 1, \\
0 & \textrm{ otherwise, }
\end{array}
\right.
\end{equation}

\noindent
where

\begin{equation}
\label{eq:ratio}
r( x, c ) = \frac{F_W(x)}{F^{C_c}_B(x)} = \sum_{i=1}^{K} \frac{ \pi^W_i \mathcal{G}(x | \mu^W_i, \Sigma^W_i) } { \pi^{C_c}_i \mathcal{G}(x | \mu^{C_c}_i, \Sigma^{C_c}_i) }.
\end{equation}

\noindent
The procedure above allows one to compute the nematode segmentation for a given image $I_t$, where $K=2$ in our system. Notice that using the ratio of MOGs (see Eq.~(\ref{eq:ratio})) is an effective way to avoid any form of thresholding. This is due to the fact that both $F_W$ and $F_B$ are explicitly modeled. Finally, opening and closing morphological operations are used to smooth nematode segmentations.

\subsection*{Nematode Skeleton}
\label{sec:meme_skeleton}

Over the years, a large number of methods have been used to extract skeletons from segmented nematodes. Methods have ranged from using specific nematode models \cite{roussel2007,huang2008}, to heuristically constructing the nematode's medial axis \cite{roussel2007, hoshi2006, baek2002, huang2009, huang2006}. While these various methods have shown success, they are generally influenced by the quality of the segmentation. In an attempt to reduce sensitivity to segmentation noise, we propose an original algorithm which balances geometric features (\ie nematode boundary) and global shape (\ie nematode undulating posture) in a seamless framework. The proposed method has the advantage of being intuitive and simple to implement.

We cast our problem once again in a probabilistic manner such that the nematode skeleton ($L_t$) is considered to be a sequence of discrete unknown skeleton pixel locations ($L_t = \{l_1,\ldots,l_M\}$), where each location is a point on the skeleton and must be determined. It is assumed here that either the head or tail pixel location ($l_0$) is initially known; determining $L_t$ is then viewed as a sequential Bayesian estimation problem \cite{kalman1960asme, isard98, sznitman2010pami}. Given the initial position $l_0$, we infer the location of the next point ($l_1$) by observing the likelihood of potential locations (\eg the likelihood of a pixel being $l_1$) as well as the history of directions between subsequent pairs of points (\eg from $l_0$ to $l_1$). The `skeleton' algorithm is iterative such that a new location along $L_t$ is inferred at each iteration step ($\tau$). To infer all points in $L_t$, this process is simply repeated.

First, the input of our algorithm is the segmentation of the nematode for a given image ($S_t$). A skeleton pixel location is defined as $l_\tau = (u_\tau,v_\tau)$, where $u_\tau$ and $v_\tau$ are pixel locations in $S_t$. Let $\nu$ be a discrete random vector describing the direction from $l_\tau$ to $l_{\tau+1}$, such that $\nu \in \mathcal{V} = \{-1,0,1\}^2$. This corresponds to $l_\tau$ being one pixel away from $l_{\tau+1}$. Let $P_\tau(\nu)$ be the corresponding probability distribution of $\nu$ at iteration step $\tau$. As more skeleton pixels are inferred, the distribution $P_\tau(\nu)$ will evolve. Initially this distribution is uniform, as no prior information between $l_\tau$ and $l_{\tau+1}$ is known. The initial position ($l_0$) is found by using maximal response locations when running a coarse corner detector on $S_t$. Selecting the following point on the skeleton can then be computed by maximizing the Maximum a Posteriori (MAP) estimator,

\begin{equation}
\label{eq:skel_map}
l_{\tau+1} = l_\tau + arg \max_{\nu} P( l_\tau | \nu ) P_\tau (\nu),
\end{equation}

\noindent
where $ P( l_\tau | \nu )$ is the likelihood that direction $\nu$ leads to the next skeleton point and is modeled by

\begin{equation}
\label{eq:skel_dist}
P( l | \nu ) = \frac {D(l+\nu)} {\sum_{\hat\nu \in \mathcal{V}} D(l+\hat\nu) }.
\end{equation}

\begin{figure}
\centering
\includegraphics[width=0.95\textwidth]{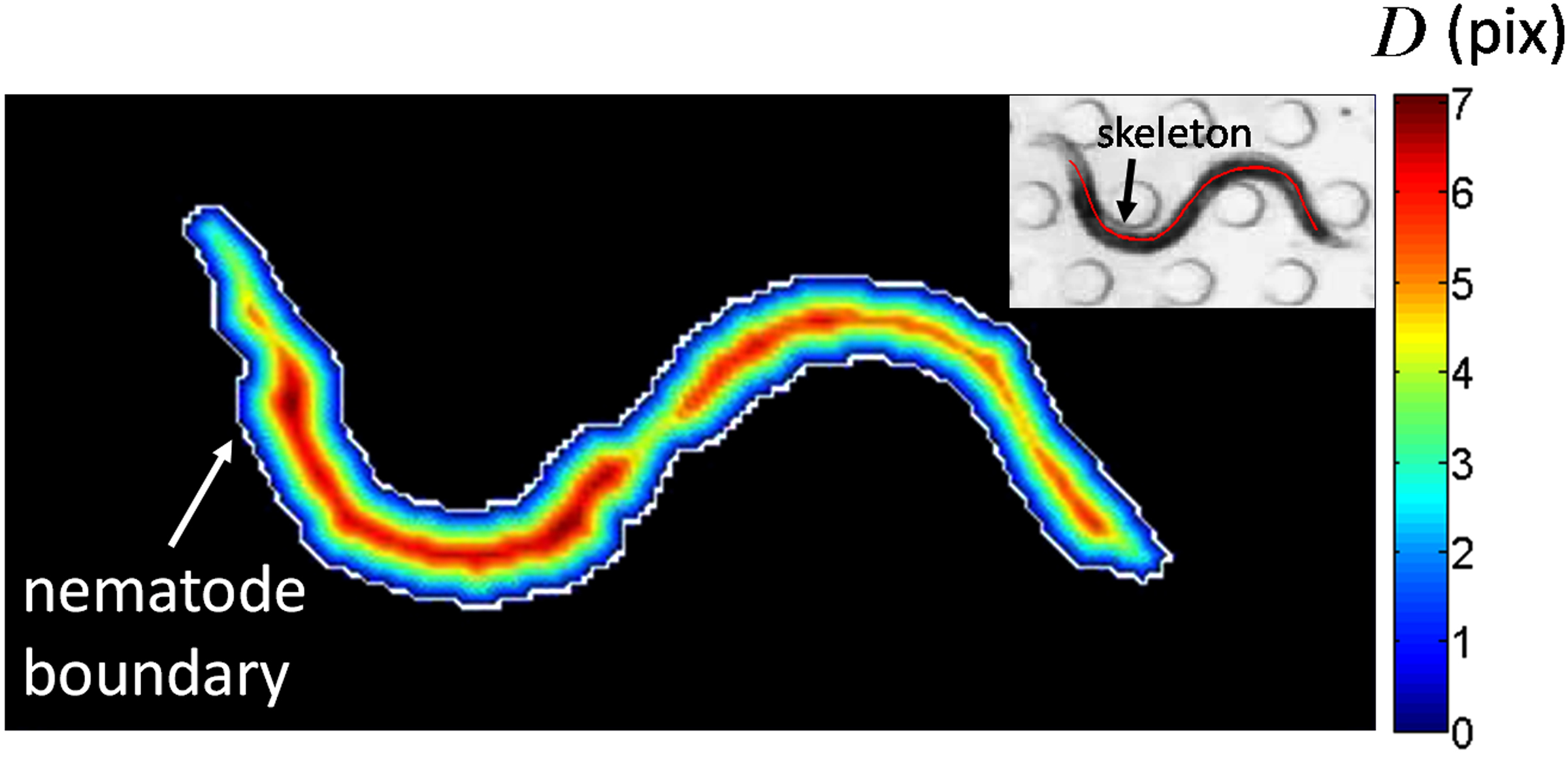}
\caption{
{\bf Computing the nematode skeleton.} Representation of the Chamfer distance transform field ($D$) applied to the segmented nematode of Fig.~\ref{Fig_4}. The value associated at each pixel of the image is the Euclidean distance (in pix) to the closest point of the nematode boundary; the distance on the boundary is zero and higher distances lie towards the nematode medial axis. {\bf (Inset)} Resulting skeleton is achieved by balancing geometric features (\ie Chamfer distance) and global shape (\ie nematode curvature).
}
\label{Fig_5}
\end{figure}

\noindent
Here, $D(l)$ is the distance computed when applying the Chamfer distance transform \cite{barrow77,gavrila07} to $S_t$. This transformation computes the Euclidean distance of each pixel in $S_t$ to its closest nematode boundary pixel. An example of this distance transform is shown in the contour plot of Fig.~\ref{Fig_5}. Here, the boundary of the nematode has a distance of zero, while values of $D$ increase steadily for pixels approaching the medial axis of the nematode. Equation~(\ref{eq:skel_map}) then implies that skeleton locations are picked by (i) weighing how likely pixels are to be at the center of the segmented nematode, combined with (ii) the history of the chosen vector directions. This strategy is particularly useful in cases where the segmentation is noisy, as the history of vector directions guides where the following pixel location should be located. In order to remove the possibility of selecting the same pixel several times, $l_\tau$ is removed from possible future locations by setting $D(l_\tau) = 0$.

Once $l_{\tau+1}$ is determined, the distribution $P_\tau(\nu)$ must be updated for the following iteration. Using Bayes rules, $P_{\tau+1}(\nu)$ is computed for $\forall \nu \in \mathcal{V}$ by

\begin{equation}
\label{eq:skel_update}
P_{\tau+1}(\nu) = \frac{1}{\mathcal{Z}} P( l_\tau | \nu ) P_\tau (\nu),
\end{equation}

\noindent
where $\mathcal{Z}$ is a normalization constant.

Inferring $L_t = \{l_1,\ldots,l_M\}$ for a given image $I_t$ then consists in the following algorithm. First, $l_0$ is given at iteration step $\tau=0$. Three steps are then repeated: (i) compute the next skeleton location from Eq.~(\ref{eq:skel_map}); (ii) update the distribution of $\nu$ from Eq.~(\ref{eq:skel_update}); and finally (iii) increment the iteration step. These operations are repeated until a point on the boundary is encountered (\ie $D(l_M) = 0$). An example of the resulting $L_t$ (skeleton) is shown in the inset of Fig.~\ref{Fig_5}.

\section*{Results}

We aim at providing a versatile nematode segmentation framework with performances comparable to, or better than, state-of-the-art image analysis systems \cite{geng2003, feng2004, cronin2005, hoshi2006, tsibidis2007, huang2009,huang2008, huang2006, baek2002}. To compare the MEME framework against such systems, we evaluate our algorithm quantitatively for a series of image sequences obtained for various {\it C. elegans} locomotive environments. The sequences include one or more data sets for behavioral assays such as (i) crawling on substrates (Supplementary Video S2 and S3), (ii) swimming in a drop (Supplementary Video S4), (iii) swimming in shallow acrylic channels (Supplementary Video S5), (iv) locomotion in a gelatin-based solution (Supplementary Video S6), and (v) locomotion in a microfluidic substrate (Supplementary Video S7). A total of 13 image sequences are investigated (see Supplementary Table S1 for complete listing and data source). In each sequence, the target nematode is present in all images. The MEME framework is implemented using Matlab; computing the nematode segmentation and skeleton for a $640 \times 480$ pixel size image requires approximately 1 second on a standard PC (\ie 2.0 Ghz).

The state-of-the-art method of choice for comparison with MEME is an in-house developed thresholding algorithm \cite{sznitman2010,sznitman2010b}, similar to standard intensity-based threshold approaches \cite{geng2003, feng2004, cronin2005, hoshi2006, tsibidis2007}. To perform a fair comparison between MEME and the thresholding framework, both methods are initially provided with a single image to tune their respective parameters. As described for MEME (see Methods), the user selects from the initial image (i) the nematode region and (ii) the nematode width (Supplementary Video S1). For the threshold-based method, all images of a sequence are first used to compute a background image of the environment by pixel averaging. Background subtraction is then applied to each image. Next, several thresholds are used to prune the remaining background pixels. These are manually selected by optimizing segmentation results on the initial image (Supplementary Table S1). Finally, opening and closing morphological operators are used to smooth and discard final background regions. Note that in the case where the number of images in the sequence is small, background subtraction is omitted and only threshold intensities are used.

\begin{figure}
\centering
\includegraphics[width=0.95\textwidth]{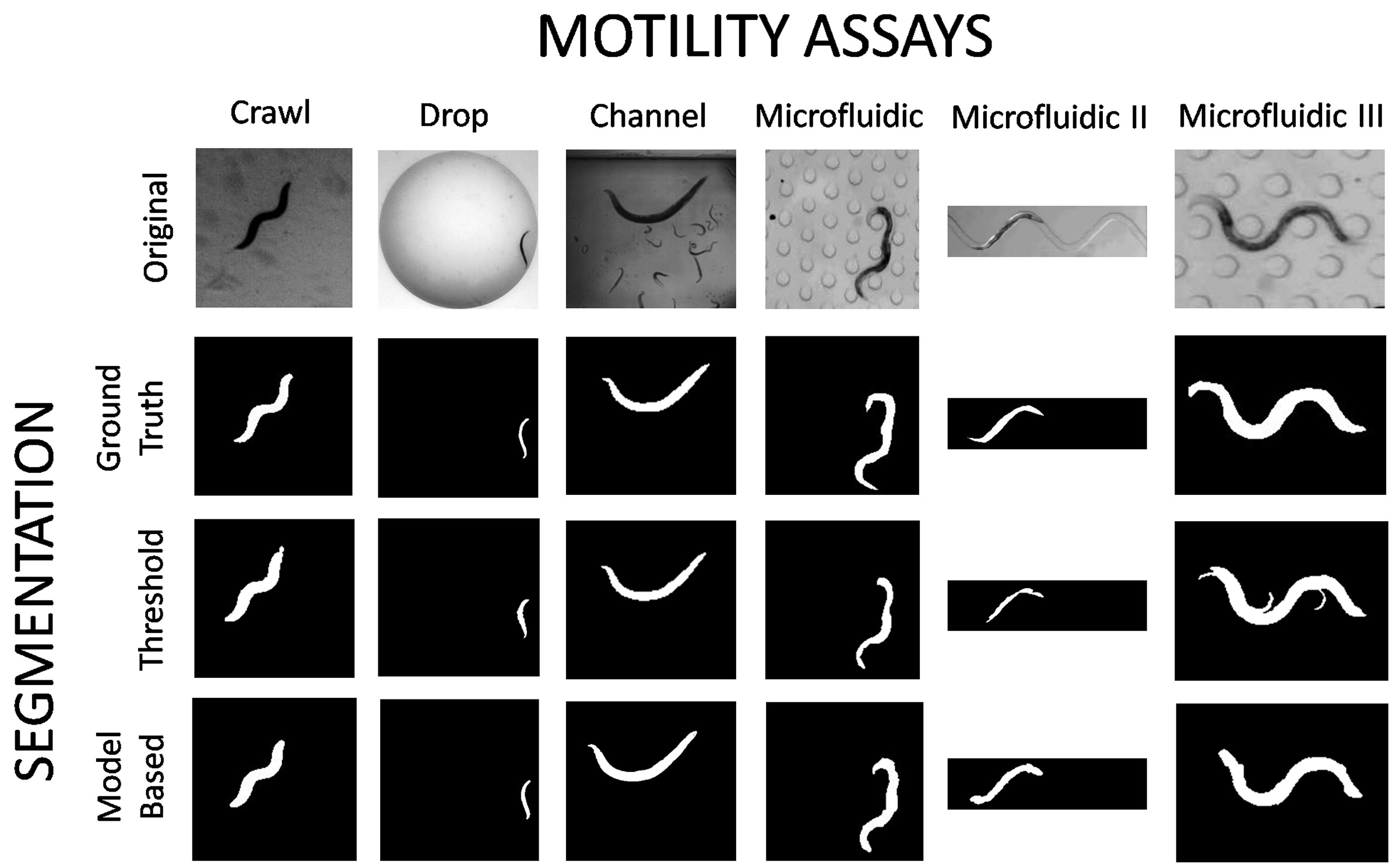}
\caption{{\bf Nematode segmentation for various locomotive environments.} {\bf (top row)} Snapshots of raw images are respectively shown for crawling (Supplementary Video S2), swimming in a drop (Supplementary Video S4), swimming in a channel (Supplementary Video S5), and locomotion in various microfluidic substrates (source: Supplemental Videos in Lockery {\it et al.} \cite{lockery2008}). Comparisons between nematode segmentations are respectively shown for (i) the ground truth, i.e. hand-segmented nematodes {\bf (second row)}, (ii) a threshold-based approach \cite{sznitman2010,sznitman2010b} {\bf (third row)}, and (iii) the Multi-Environment Model Estimation (MEME) algorithm {\bf (bottom row)}. See Supplementary Table S1 for data on all 13 cases investigated.}
\label{Fig_6}
\end{figure}

In order to quantitatively evaluate any segmentation algorithm, results must be compared to a \textit{ground truth} \cite{sznitman2009iccv}. For the present purpose, the ground truth is set as the true, or optimal, nematode segmentation provided by an expert. Hence, we manually segment a small set of images ($n=30$~-~40) from each sequence (\eg Fig.~\ref{Fig_6}, second row) and compare the performance of each algorithm to this image sub-set. Determining such ground truth allows for a precise definition of correct and incorrect pixel classification.

The performance of a segmentation algorithm can be evaluated by measuring two distinct metrics \cite{sznitman2009iccv}: (i) the {\it surface error} and (ii) the {\it nematode yield}. The former quantity computes the proportion of pixels which are misclassified by the algorithm over the entire image. This metric is mathematically defined as

\begin{equation}
\label{eq:surferr}
\epsilon_{t} = \frac{1}{|S_t|}\sum_{\forall p \in S_t} | G_t(p) - S_t(p) |,
\end{equation}

\noindent
where $p$ is a pixel location and $G_t$ is the ground truth image for $I_t$. Hence, $\epsilon_{t}$ attributes equivalent weight to errors on the background and the nematode regions. The nematode yield, however, only computes the proportion of the nematode region which is correctly segmented. Consequently, misclassified pixels belonging to the background have no impact on the nematode yield. This metric is defined as

\begin{equation}
\label{eq:nemyield}
\eta_{t} = \frac{1}{|\mathcal{N}|}\sum_{\forall p \in \mathcal{N}} | G_t(p) - S_t(p) |,
\end{equation}

\noindent
where $\mathcal{N}$ is the set of pixels which satisfy $G_t(p) = 1$. Together, $\epsilon_{t}$ and $\eta_{t}$ provide a quantitative and reliable measure of segmentation performance \cite{sznitman2009iccv}.

Qualitative segmentation results are shown for a selection of motility environments in Fig.~\ref{Fig_6} as well as in the Supplementary Videos S2 to S7. In general, the MEME method is capable of segmenting nematodes at least as well as the thresholding method. For the ``Crawl'', ``Drop'', and ``Channel'' environments (first three columns, Fig.~\ref{Fig_6}), both methods yield qualitatively similar results. Here, the environments illustrate a relatively homogenous background. However, in the complex ``Microfluidic'' environments (Fig.~\ref{Fig_6}), results contrast more sharply between the two approaches; MEME provides cleaner segmentations which capture more closely the original shape of the nematodes.

\begin{figure}
\centering
\includegraphics[width=0.95\textwidth]{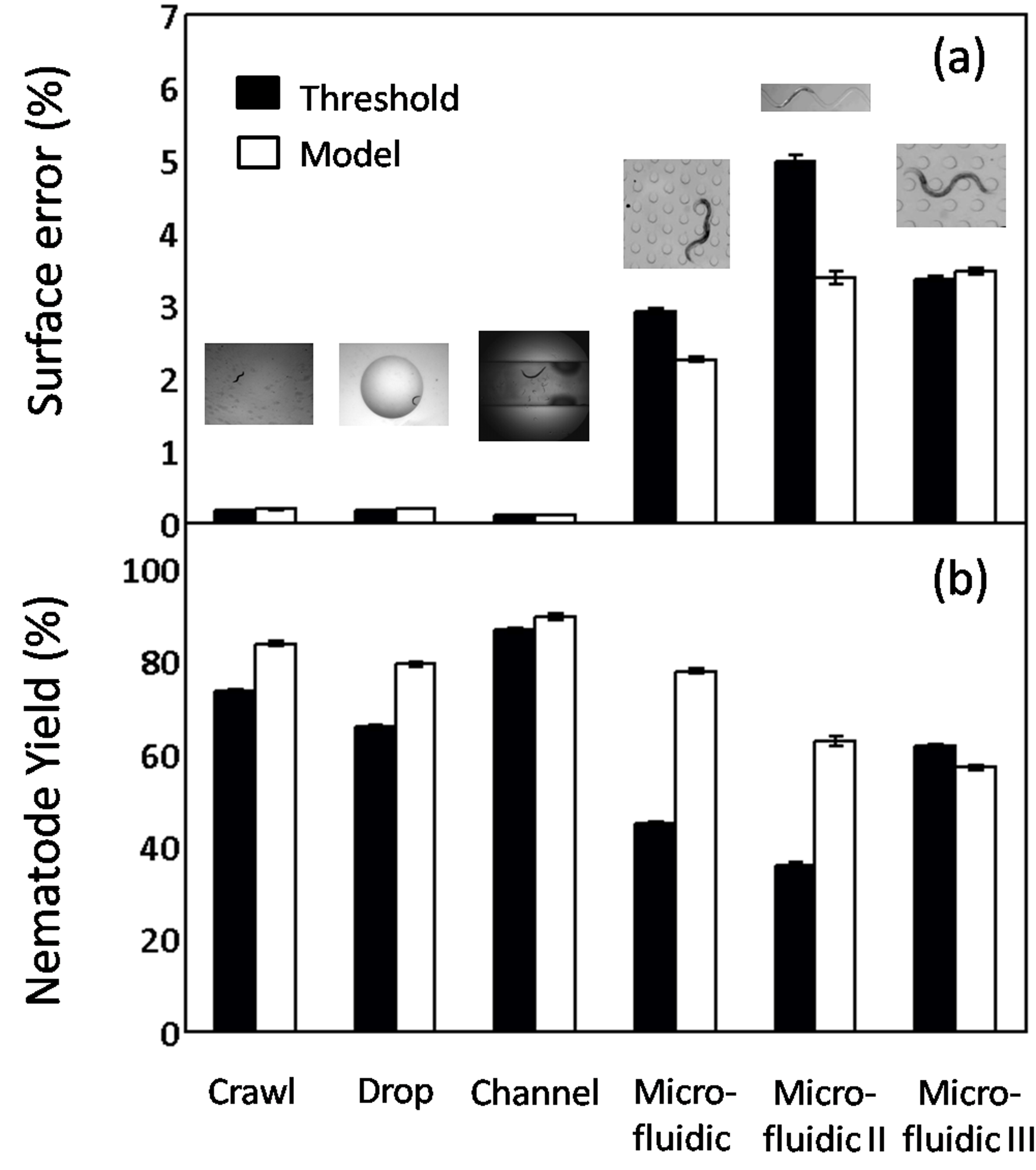}
\caption{{\bf Performance evaluation of nematode segmentation algorithms}. Here, the Multi-Environment Model Estimation (MEME) algorithm is compared to a state-of-the-art thresholding approach \cite{sznitman2010,sznitman2010b} for the environments shown in Fig.~\ref{Fig_6}. {\bf (a)} Surface error ($\epsilon_{t}$): proportion of pixels which are misclassified by an algorithm over the entire image (see Eq.~(\ref{eq:surferr})). {\bf (b)} Nematode yield ($\eta_{t}$): proportion of the nematode region which is correctly segmented (see Eq.~(\ref{eq:nemyield})). Complete data on surface error and nematode yield is available in the Supplementary Table S1 for all 13 cases investigated.}
\label{Fig_7}
\end{figure}

Figure~\ref{Fig_7} shows the results from the computation of the surface error (Fig.~\ref{Fig_7}a) and the nematode yield (Fig.~\ref{Fig_7}b) for the environments of Fig.~\ref{Fig_6}. Data for $\epsilon_{t}$ and $\eta_{t}$ is reported in the Supplementary Table S1 for the complete 13 image sequences. In general, computations of surface error ($\epsilon_{t}$) illustrate comparable performances between MEME and the threshold-based approach (Fig.~\ref{Fig_7}a). Yet, in two complex ``Microfluidic'' environments, MEME does significantly better. Note that for all environments investigated here (Fig.~\ref{Fig_7}a and Supplementary Table S1), $\epsilon_{t}$ remains below $10\%$. In fact, for homogeneous background environments such as ``Crawl'', ``Drop'', and ``Channel'', $\epsilon_{t} \ll 1\%$, emphasizing the good results obtained both by MEME and the threshold-based approach.

We observe, in contrast, significant improvements in nematode yield ($\eta_{t}$) when using MEME compared to the thresholding method (Fig.~\ref{Fig_7}b). From the set of 13 assays tested here, 10 cases show examples of MEME significantly outperforming the threshold-based method (Supplementary Table S1); in some cases, with margins greater than 20 percentage points (\eg ``Microfluidic'' and ``Microfluidic II'', Fig.~\ref{Fig_7}b). In the remaining environments where the thresholding method performs relatively better (\eg ``Microfluidic III'', Fig.~\ref{Fig_7}b), the differences in $\eta_{t}$ remain however small, \ie between 1.79 and 4.71 percentage points. Overall, our MEME algorithm outperforms the threshold-based approach for the overwhelming majority of cases examined.

\section*{Discussion}

Our experiments show that MEME provides a reliable framework to obtain nematode segmentations of {\it C. elegans} across various locomotive environments. In addition, MEME offers significant improvements over alternative image analysis systems available; these include (i) better, or similar, performances compared to state-of-the-art thresholding approaches \cite{sznitman2010,sznitman2010b}, (ii) no nematode-free image sequence required for learning background appearances \cite{roussel2007, tsechpenakis2008}; and (iii) a small amount of user input needed, \ie a single hand-segmentation of the nematode and a marking of its width (Supplementary Video S1). This last improvement is particularly attractive from a user point of view as substantial effort may be needed with thresholding techniques to obtain similar results. Overall, these attributes make the MEME framework both attractive and straightforward to use for a broad range of researchers.

While computing good nematode segmentations with threshold-based methods is possible (Fig.~\ref{Fig_7}), this process can quickly become laborious. Indeed, several iterations are required by the user to find optimal thresholds for a given environment (Supplementary Table S1). The main complication arises from the non-uniform backgrounds and appearance (\ie pixel intensities) which characterize many environments. For example, a single threshold is incapable of distinguishing between the nematode and the background in the presence of pillars in microfluidic substrates (\eg Fig.~\ref{Fig_6}, last column). Similarly, single thresholds cannot adapt to specific locations in an image. This becomes crucial for accurate segmentation of nematodes in environments where lighting conditions may not be uniform (\eg Supplementary Video S5).

In general, the improvement observed with MEME can be attributed to two main reasons: (1) the nematode appearance model is explicitly learned and used to help decide whether pixels belong to the nematode. In practice, when using threshold-based methods, many of the regions which are considered ``not background" after applying a threshold do not resemble the nematode at all (\eg pillars in ``artificial dirt'' assays of Lockery {\it et al.} \cite{lockery2008}, Fig.~\ref{Fig_1}f and Supplementary Video S7). Using both the nematode and background appearance models significantly reduces the need of using intense pruning schemes to reject such regions. (2) The background scene is partitioned into a grid of sub-regions (Fig.~\ref{Fig_4}c), where each cell is explicitly modeled. This allows for local variations in intensities to be grouped by region, providing a localized statistical model for each area of the background scene. This strategy has the advantage of appropriately modeling backgrounds where large variations in lighting occur (\eg Fig.~\ref{Fig_1}g).

In cases where the nematode appearance differs significantly from the background, such as in crawling and swimming assays (\eg Fig.~\ref{Fig_1}a and b), we observe nematode yields ($\eta_t$) beyond $80\%$ (Fig.~\ref{Fig_7} and Supplementary Table S1). In contrast, more complex environments can substantially reduce this performance (\eg microfluidic substrates). The main difficulty therein lies in that only pixel intensities are modeled; this represents an important limitation when pixel intensities of the nematode and the background are too similar. A typical illustration of this problem occurs at the head and tail of {\it C. elegans}, where the nematode extremities are often transparent against the background. For example, this problem is observed in microfluidic substrates (\eg Fig.~\ref{Fig_1}e and f) where the ends of the nematode are lost during the segmentation process. A direct consequence of this is the truncated length of nematode skeletons (\eg inset of Fig~\ref{Fig_5} and Supplementary Videos S7).

Our MEME framework is currently optimized for segmenting a single target nematode within an image sequence. Nevertheless, scenarios where multiple nematodes enter the scene in subsequent images are still supported by our algorithm as long as only one nematode is present in the input image. That is, an arbitrary number of nematodes may be segmented for a given image sequence. Note, however, that cases where the appearance of either the nematode or the background changes significantly over the length of an image sequence will cause improper segmentations. Furthermore, extracting skeletons remains a challenge in some scenarios. For example, cases where the nematode coils on itself, or when its head and tail touch (\eg omega bend), are currently not supported with the implemented skeleton algorithm. In the former case, the segmentation simply does not provide a correct shape representation of the nematode (\ie a closed circle as opposed to a `snake'). The problem lies in the fact that estimating the medial axis of the nematode with the Chamfer distance transform is ill-suited. In principle, the latter scenario (\ie omega bend) is not problematic. In practice, however, initializing the skeleton algorithm is ill-posed; there is no a priori knowledge as to where the head or tail lie.

\subsection*{Motility Metrics}

\begin{figure}[!ht]
\centering
\includegraphics[width=1.0\textwidth]{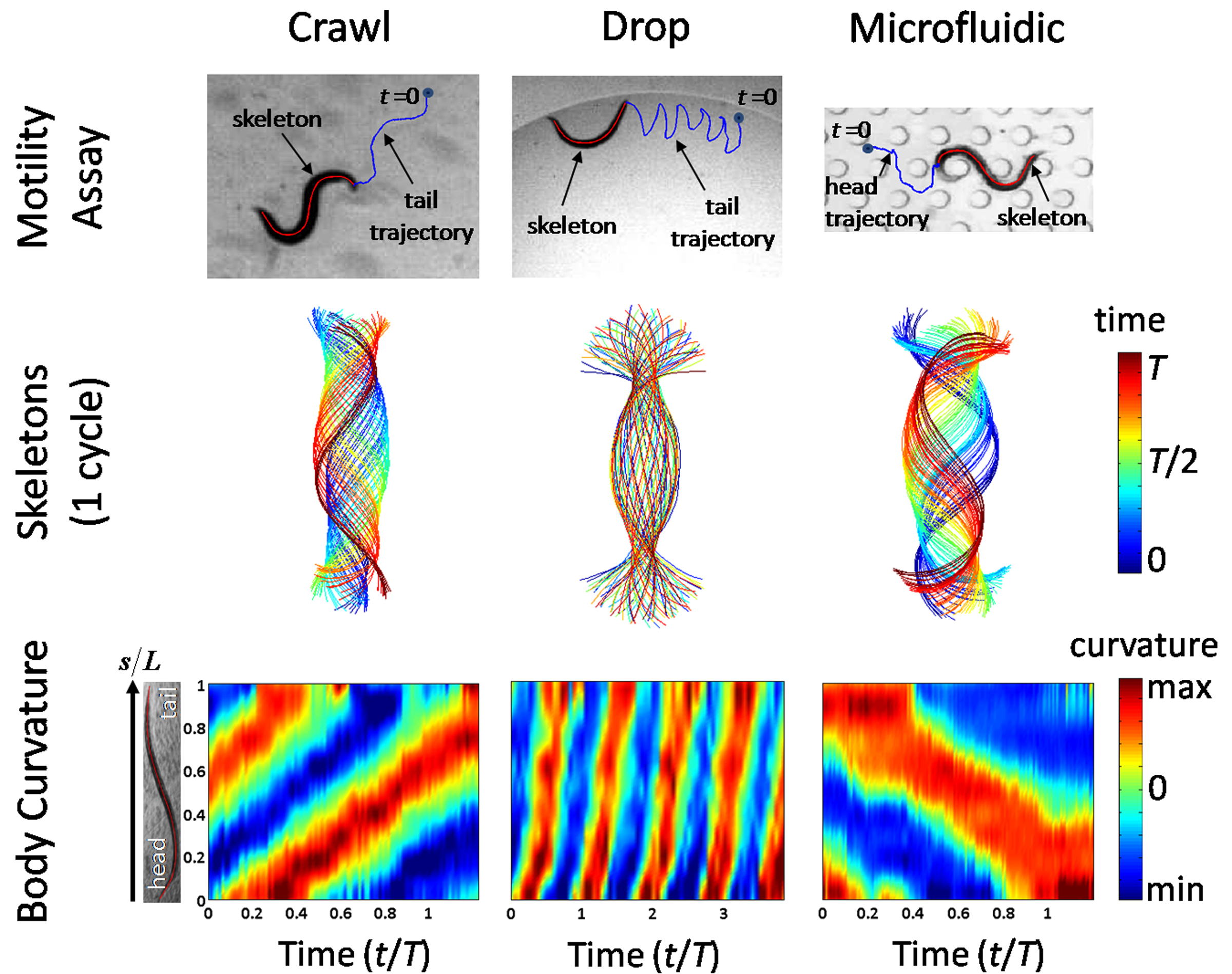}
\caption{{\bf Examples of nematode locomotive features in sample environments.} Here, nematode skeleton data is shown for a crawling assay (left column), for swimming in a 5~ml drop (middle column), and for locomotion in a microfluidic substrate obtained from Lockery {\it et al.} \cite{lockery2008} (right column). Additional skeleton data is shown in Supplementary Videos S2 to S7. {\bf (top row)} Tracking data of path swept by nematode head (or tail) over multiple beating cycles. {\bf (middle row)} Color-coded temporal evolution of \emph{C. elegans} skeletons over one beating period ($T$). Results reveal a distinct envelope of body postures for each environment. {\bf (bottom row)} Spatio-temporal contour plot of body curvature ($\kappa$) along the length of the nematode's skeleton. Red and blue colors represent positive and negative curvature  values, respectively. The $y$-axis corresponds to the dimensionless position ($s/L$) along the \emph{C. elegans}' body length where $s=0$ is the head and $s=L$ is the tail.}
\label{Fig_8}
\end{figure}

We briefly discuss the feasibility of using nematode skeletons obtained with MEME (Supplementary Videos S2 to S7). Skeleton data often provides the building blocks to quantify locomotive traits of {\it C. elegans}~\cite{karbow2006,korta2007,pierce2008,sznitman2010,sznitman2010b}. Here, we illustrate some of these motility metrics across sample environments. In Fig.~\ref{Fig_8} (top row), nematode tracking data is shown over multiple body bending cycles for crawling on a substrate (left column), swimming in a drop (middle column), and locomotion in so-called ``artificial dirt'' (right column), \ie a microfluidic substrate (Supplemental Videos in Lockery {\it et al.} \cite{lockery2008}). Trajectories swept by the nematode tail (or head) are labeled, illustrating striking differences in the travel paths adopted by {\it C. elegans} as a function of the surrounding environment. Snapshots of nematode skeletons over one beating cycle are shown in Fig.~\ref{Fig_8} (middle row); the time evolution of skeletons is color-coded as a function of the corresponding beating period ($T$). Plots reveal the existence of well-confined envelopes of body postures which vary dramatically with motility assay. Here, envelopes of postures are constructed using a principal component analysis (PCA) to find the skeleton's principal axis and orientation at each instant in time. Further metrics including the nematode wavelength as well as the amplitude of body undulations can be obtained in a straightforward manner from the construction of such envelopes \cite{sznitman2010}.

Next, we illustrate measures of body curvature ($\kappa$) along the nematode's length (Fig.~\ref{Fig_8}, bottom row); such plots have been shown to characterize swimming and crawling gaits \cite{korta2007,pierce2008,sznitman2010}. Curvature is defined as $\kappa(s,t)=d \phi / ds$, where $\phi$ is the angle made by the tangent to the $x$-axis at each point along the nematode skeleton; $s$ is the arc-length coordinate spanning the nematode's head ($s=0$) to its tail ($s=L$). The spatio-temporal evolution of $\kappa$ is shown over several beating cycles for each environment. Here, curvature values are color-coded; red and blue represent positive and negative values of $\kappa$, respectively. Note that the vertical axis in each contour plot corresponds to the non-dimensional body position ($s/L$), where $L$ is the nematode length. Each contour plot shows the existence of highly periodic, well-defined diagonally oriented lines. These diagonal lines are characteristic of bending waves of motion which propagate in time along the nematode body length (\ie traveling waves).

In Fig.~\ref{Fig_8}, forward motion displays curvature lines with a positive slope (left and middle column); waves are initiated at the nematode head \cite{pierce2008,sznitman2010}. Conversely, backward motion displays lines with a negative slope, where bending motion is initiated at the tail (right column). In general, a number of motility metrics may be directly extracted from such curvature contour plots. For example, the body bending frequency ($f$) may be obtained by applying a one-dimensional (1D) Fast Fourier Transform (FFT) to the curvature field $\kappa$ at multiple body positions $s/L$ \cite{sznitman2010}. Similarly, the wave speed ($c$) may be directly extracted from the slope of the curvature $\kappa$ propagating along the nematode's body; the wavelength $\lambda =c/f$ is then computed in a straightforward manner. With our MEME platform, nematode skeleton data is made ready available for use for motility analysis of {\it C. elegans}.

\subsection*{Future Directions}

The proposed MEME framework provides researchers with an attractive and reliable platform for nematode segmentation and `skeletonizing' across a large spectrum of {\it C. elegans} motility assays. The MEME software is freely available upon request (contact person: J. Sznitman; website for download will be provided). Improving our system to further assist researchers conduct quantitative analysis of {\it C. elegans} is of course desired. In the near future, one immediate goal is to provide segmentations and skeletons simultaneously for multiple nematodes. This `upgrade' would be of great interest for high-throughput drug screening applications \cite{rhode2007,shi2008}. From a performance point of view, combining larger sets of image features (\eg edges, texture, etc.) with MOG models may provide better appearance models for difficult environments. This may yield better segmentation results, in particular at the nematode extremities (\ie head and tail). Finally, our MEME platform is not restricted to image analysis of {\it C. elegans} only. For instance, MOG methods may be used for applications relating cell tracking and motility \cite{dublin2008,li2008,xiong2006}. We illustrate here an example of such possible application with Albino Swiss mouse embryo fibroblast cells (Supplementary Video S8).


\section*{Supporting Information}
{\bf Table S1:} Compiled data on segmentation results for the Multi-Environment Model Estimation (MEME) and threshold-based algorithms. Performances of each algorithm (\ie surface error and nematode yield) are evaluated for 13 different image sequences representative of various locomotive environments (\eg crawling on agar plate, swimming in a channel or a drop, locomotion in microfluidic substrates).

~\\
{\bf Video S1:} MEME software tutorial shown for a sample image sequence (source: berrigel2.0perc.mov, Supplementary Material in Berri {\it et al.} \cite{berri2009}).

~\\
{\bf Video S2:} Example of crawling assay. From left to right: (i) raw image, (ii) binary segmentation from MEME, and (iii) resulting skeleton superimposed on raw image. Here, a young adult, wild-type (N2) {\it C. elegans} is seen crawling on an agar plate. Nematode is approximately 1~mm long (image resolution: $1/78~$mm/pix; image acquisition rate: 28 frames per second).

~\\
{\bf Video S3:} Example of crawling on a substrate. From left to right: (i) raw image, (ii) binary segmentation from MEME, and (iii) resulting skeleton superimposed on raw image. The original data is obtained from the Supplementary Information (Movie 1) in Pierce-Shimomura {\it et al.} \cite{pierce2008}.

~\\
{\bf Video S4:} Example of swimming assay in a liquid drop. From left to right: (i) raw image, (ii) binary segmentation from MEME, and (iii) resulting skeleton superimposed on raw image. Here, a young adult, wild-type (N2) {\it C. elegans} is seen swimming in a 5~ml drop of M9 buffer solution. Nematode is approximately 1~mm long (image resolution: $1/78~$mm/pix; image acquisition rate: 28 frames per second).

~\\
{\bf Video S5:} Example of swimming assay in a shallow channel. From left to right: (i) raw image, (ii) binary segmentation from MEME, and (iii) resulting skeleton superimposed on raw image. Here, a young adult, wild-type (N2) {\it C. elegans} is seen swimming in a narrow acrylic channel filled with M9 buffer solution. Details are given in Sznitman {\it et al.} \cite{sznitman2010}. Nematode is approximately 1~mm long (image resolution: $1/400~$mm/pix; image acquisition rate: 125 frames per second).

~\\
{\bf Video S6:} Example of nematode locomotion in a gelatin-based solution. From left to right: (i) raw image, (ii) binary segmentation from MEME, and (iii) resulting skeleton superimposed on raw image. The original data is obtained from the Supplementary Information (berrigel2.0perc.mov) in Berri {\it et al.} \cite{berri2009}.

~\\
{\bf Video S7:} Example of nematode locomotion in ``artificial dirt'', \ie a microfluidic substrate. From left to right: (i) raw image, (ii) binary segmentation from MEME, and (iii) resulting skeleton superimposed on raw image. The original data is obtained from the Supplementary Information (Video 2) in Lockery {\it et al.} \cite{lockery2008}.

~\\
{\bf Video S8:} Application of the MEME software to segmentation of Albino Swiss Mouse Embryo Fibroblast cells (3T3 Line). The original video (Video 2) is extracted from live-cell imaging videos available at {\it Nikon Microscopy U} (http://www.microscopyu.com/moviegallery/livecellimaging/3t3/index.html), as obtained with Differential Interference Contrast (DIC) microscopy.

\section*{Acknowledgments}
Strains of {\it C. elegans} were obtained from the Caenorhabditis elegans Genetic Stock
Center (University of Minnesota), supported by the National Institutes of Health (Bethesda MD, USA). The authors would like to thank Dr. R. Ghosh (Lewis-Sigler Institute for Integrative Genomics, Princeton U.), X. Shen (Department of Mechanical Engineering and Applied Mechanics, U. of Penn), and Dr. A.E.X. Brown (MRC Lab of Molecular Biology, Cambridge U.) for their support and helpful discussions.

\section*{Author Contributions}
Conceived and designed the experiments: RS JS. Performed the experiments: RS MG. Analyzed the data: RS MG. Contributed reagents/materials/analysis tools: RS MG GDH PEA JS. Wrote the paper: RS JS.

\bibliographystyle{plain}
\bibliography{memebib}

\section*{Figure Legends}

\end{document}